\documentclass[10pt,twocolumn,letterpaper]{article}

\usepackage{icb}

\usepackage{times}
\usepackage{epsfig}
\usepackage{graphicx}
\usepackage{amsmath}
\usepackage{amssymb}
\usepackage{float}
\usepackage{multirow}
\makeatletter
\newcommand*{\rom}[1]{\expandafter\@slowromancap\romannumeral #1@}
\makeatother

\usepackage[norule,symbol,perpage]{footmisc}

\icbfinalcopy 

\begin{document}

\title{ID Preserving Generative Adversarial Network for Partial Latent Fingerprint Reconstruction}

\author{Ali Dabouei, Sobhan Soleymani, Hadi Kazemi, Seyed Mehdi Iranmanesh, Jeremy Dawson,\\ Nasser M. Nasrabadi, {\it Fellow, IEEE }, West Virginia University\\
{\tt\small \{ad0046, ssoleyma, hakazemi, seiranmanesh\}@mix.wvu.edu,}
\\
{\tt\small \{jeremy.dawson, nasser.nasrabadi \}@mail.wvu.edu}
}

\maketitle

\thispagestyle{empty}

\begin{abstract}
Performing recognition tasks using latent fingerprint samples is often challenging for automated identification systems due to poor quality, distortion, and partially missing information from the input samples. We propose a  direct latent fingerprint reconstruction model based on conditional generative adversarial networks (cGANs).
Two modifications are applied to the cGAN to adapt it for the task of latent fingerprint reconstruction. First, the model is forced to generate three additional maps to the ridge map to ensure that the orientation and frequency information are considered in the generation process, and prevent the model from filling large missing areas and generating erroneous minutiae.
Second, a perceptual ID preservation approach is developed to force the generator to preserve the ID information during the reconstruction process. 
Using a synthetically generated database of latent fingerprints, the deep network learns to predict missing information from the input latent samples.
We evaluate the proposed method in combination with two different fingerprint matching algorithms on several publicly available latent fingerprint datasets. We achieved rank-10 accuracy of 88.02\% on the IIIT-Delhi latent fingerprint database for the task of latent-to-latent matching and rank-50 accuracy of 70.89\% on the IIIT-Delhi MOLF database for the task of latent-to-sensor matching. 
Experimental results of matching reconstructed samples in both latent-to-sensor and latent-to-latent frameworks indicate that the proposed method significantly increases the matching accuracy of the fingerprint recognition systems for the latent samples.  
\end{abstract}

\section{Introduction}

Automatic fingerprint recognition systems have been widely adopted to perform reliable and highly accurate biometric identification. Compared to other biometric traits, such as iris, the fingerprint has a unique superiority of being collected indirectly from crime scenes from latent friction ridge impressions.
Fingerprint samples can be categorized into three main groups based on the acquisition techniques such as: inked, live-scan, or latent samples. The inked and live-scan samples are considered as clean samples for which users leave impressions intentionally in access control or authentication scenarios. In addition, an agent, or the acquisition process of the system itself, can monitor quality of the samples, and guide users to leave appropriate fingerprints. In past decades, algorithms for preprocessing and matching clean fingerprints have advanced rapidly, resulting in the development of numerous and varied commercial fingerprint recognition systems.

\begin{figure}
\begin{center}
\includegraphics[scale=.19]{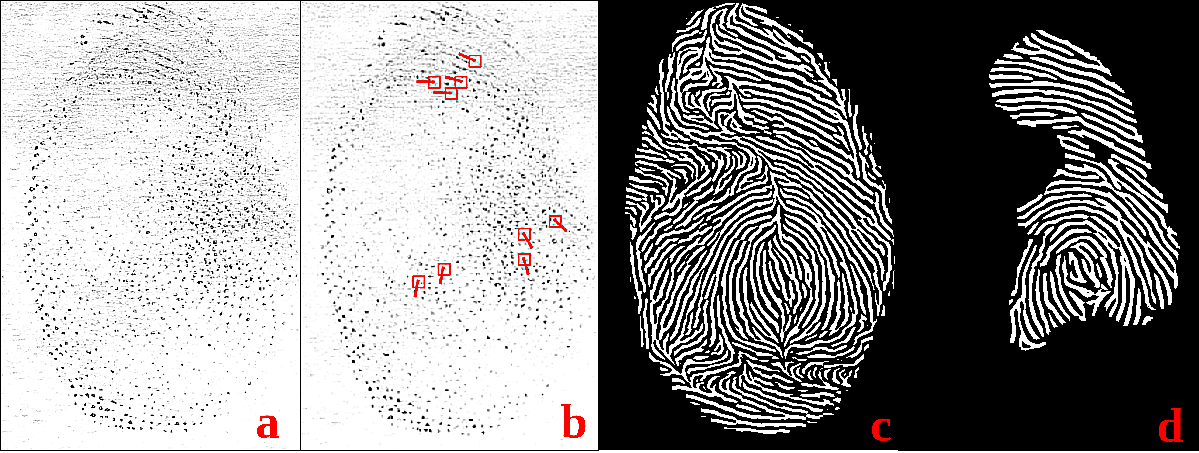}
\end{center}
  \caption{Examples of different latent fingerprint reconstruction methods: a) a latent fingerprint with severe distortion and missing area, b) minutiae-based prediction using \cite{sankaran2017group},  c) ridge-based reconstruction using \cite{svoboda2017}, d) constrained ridge-based reconstruction using the proposed algorithm. }
\label{fig:samples}
\end{figure}

In contrast, latent fingerprints are the marks of fingers unintentionally left on the surface of an object in a crime scene. Typically, a latent fingerprint is a `noisy' image with a notable missing area, therefore containing a lower amount of ridge information (i.e. minutiae) compared to inked or live-scan fingerprints. Processing latent fingerprints is a complex and challenging problem due to the under-determined properties of the problem, and presence of many disturbing factors introduced by background objects or patterns on the substrate or surface, the force and torque involved in depositing the latent print, etc. Commercial and state-of-the-art methods for recognizing the inked or live-scan fingerprints often fail to process latent samples, even in the preprocessing stage \cite{maio2004fvc2004}. Therefore, various approaches have been proposed in the literature to tackle the problem of latent \cite{li2018deep, yoon2011latent, yoon2010latent, feng2013orientation, cao2015latent} and distorted \cite{dabouei2018fingerprint, DetectionandrectificationSi, ref2} fingerprints.

Enhancing latent fingerprints often leads to optimizing a cost function that measures the quality of reconstruction by comparing reconstructed information and their ground truths. Based on the type of the reconstructed information, latent fingerprint reconstruction methods can be divided into two main categories: ridge-base and minutiae-based methods. 

In the ridge-base methods \cite{svoboda2017, yoon2012, yoon2010latent, yoon2011latent, feng2013orientation}, algorithms try to predict the ridge information, which can be the orientation map or the ridge pattern itself, and minimize the similarity between the generated information and their ground truths. Then minutiae information can be extracted from the predicted ridge information. These methods are optimized to predict the ridge information without estimating the local quality of the input samples. For parts of the input latent fingerprint that there exists some information, typically these algorithms produce useful results. However, for the parts in which there is a severe distortion, not only these algorithms can not predict the missing information, but also, they can destroy the ID information by generating erroneous minutiae. Figure \ref{fig:samples}(c) shows a reconstructed ridge map for a latent fingerprint with severe distortion. As the reconstructed ridge map indicates, algorithm \cite{svoboda2017} generates meaningful ridge information for parts that contain some ridge information in the input latent sample, but for other parts it produces random ridge patterns which drastically decreases the matching score.

On the other hand, minutiae-based methods \cite{tang2017latent, sankaran2017group, sankaran2014latentstacked} directly predict the type and location of minutiae of the latent fingerprints without reconstructing the ridge pattern. Minutiae-based reconstruction methods are more robust against severe distortion or missing areas of the input latent fingerprint, since they predict the probability of a minutia by analyzing a small area around each candidate ridge point, and they reject large missing areas. Hence, this rejection drastically decreases the number of founded minutiae. Figure \ref{fig:samples}(b) shows some minutiae that were detected using local processing of a latent fingerprint.   

In case of severely distorted latent fingerprints, both ridge-based and minutiae-based reconstruction methods fail. Ridge-based reconstruction methods fail because they fill missing areas with incorrect ridge patterns, therefore they introduce minutiae which change the ID of reconstructed samples. Minutiae-based prediction methods fail because they reject most of the missing areas, and at the end, they often predict fewer minutiae, which are not enough to identify samples.         

In this study we developed a deep convolutional neural network (DCNN) model to reconstruct the ridge information of latent fingerprints. The core network in the model is a conditional generative adversarial network (cGAN) that reconstructs the obscured ridge information of the latent samples. To overcome the limitation of the previous ridge-based reconstruction methods, our model predicts three extra maps in addition to the ridge map: the orientation, frequency and segmentation maps. Generating the orientation and frequency maps ensure that the model is considering the orientation and frequency information of the input latent fingerprints. Generating a segmentation map prevents the model from filling large missing areas in the input latent samples; thus, it optimizes the amount of ridge information that can be reconstructed. In addition, to force the generator to preserve the ID information (type and location of minutiae), we developed an auxiliary deep model to extract the perceptual ID information (PIDI) of the generated sample and fuse it into the cGAN model to enhance the reconstruction process.

\begin{figure*}[t]
\begin{center}
\includegraphics[scale=.30]{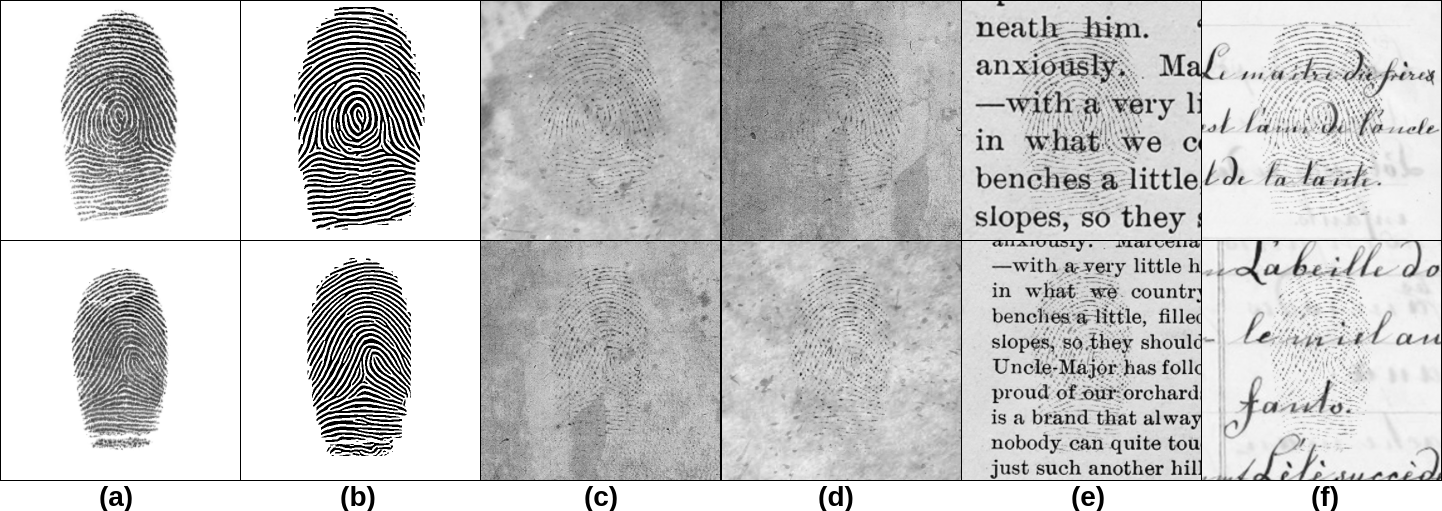}
\end{center}
   \caption{Examples of synthetically generated latent fingerprints: (a) is the original fingerprint, (b) is the corresponding binary ridge map, and (c, d, e, f) are generated latent fingerprints.}
\label{fig:latent}
\end{figure*}

\begin{figure}
\begin{center}
\includegraphics[scale=.52]{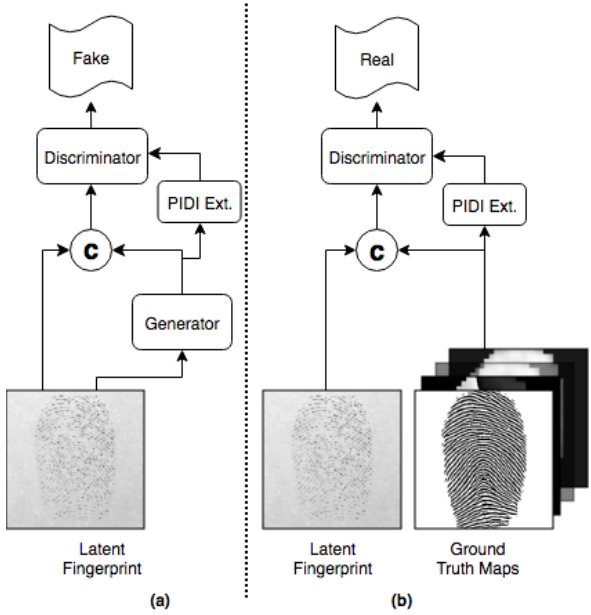}
\end{center}
  \caption{Training criteria of the discriminator. The discriminator receives two types of inputs, (a) the generated fingerprint maps, and (b) the ground truth maps. In both scenarios, the corresponding latent fingerprint is concatenated (\textbf{C}) to either the generated or the ground truth maps to act as the condition. The auxiliary verifier module extracts the PIDI from both generated maps and ground truths, and passes them to the discriminator. The discriminator learns to distinguish between the real maps and fake maps based on the quality and PIDI of generated maps.}
\label{fig:discrim}
\end{figure}

\section{Method}
\label{Method}

\subsection{Conditional Generative Adversarial Networks (cGANs)}
GANs \cite{goodfellow2014generative} are one of the most popular groups of generative networks. Generative networks map a sample from a random distribution $p_z(z)$ to a target domain of desired samples $y = G(z,\theta_g) : z \rightarrow y$, through training parameters $(\theta_g)$ of the network. GANs are different from conventional generative models because they prosper from a discriminator network. The discriminator network compares the generated samples (fake samples) $y = G(z,\theta_g)$ with the real samples from the target domain, and tries to distinguish between them. Simultaneously, the generator (typically an auto-encoder) tries to fool the discriminator by generating more realistic samples. In each iteration, the generator produces better samples in an attempt to fool the discriminator, and the discriminator improves by comparing the real samples with the generated samples. In other words, the discriminator $D$ and the generator $G$ play a two-player minimax game with the following objective function:
\begin{align}
\label{eq_gan}
V_{GAN}(G,D) = &E_{y\sim P_{data}(y)}[log D(y)] +\\ \nonumber &E_{z\sim P_{z}(z)}[log (1-D(G(z)))],
\end{align}
where generator $G$ tries to minimize Eq. \ref{eq_gan} and discriminator $D$ tries to maximize it. An additional $L2$ or $L1$ distance loss is added to the objective function in the literature to force the network to generate samples which are closer to the target ground truth. The final generator model trains as follows:

\begin{align}
\label{eq_train}
G_{optimal} = \min_G\max_DV_{GAN}(G,D) + \lambda l_{L1}(y^*, G),
\end{align}
where $\lambda$ is the coefficient of $L1$ distance and $l_{L1}(y^*, G)$ is:

\begin{align}
\label{eq_l1}
l_{L1}(G) = ||y^*-G(z)||_1,
\end{align}
where $y^*$ is the ground truth for the output of the generator. 

In the real-world application of restoring a latent fingerprint image, both the source and target domains are available for training. Therefore, Isola et al. \cite{isola2016image} proposed the conditional GAN model which has two modifications compared to conventional GANs. First, instead of using noise as the input to the network, real training samples from the source domain are fed to the network $y = G(x,\theta_g) : x \rightarrow y$. Second, they put a condition on the discriminator by concatenating the input sample with the generated sample. Therefore, the discriminator judges a generated sample based (conditioned) on the original sample which was fed to the network. The new objective function for the cGAN is:
\begin{align}
\label{eq_cgan}
V_{cGAN}(G,D) = &E_{x\sim P_{data}}[log D(x,y)] +\\ \nonumber &E_{x\sim P_{data}}[log (1-D(x,G(x)))].
\end{align}

\begin{figure*}[t]
\begin{center}
\includegraphics[scale=.32]{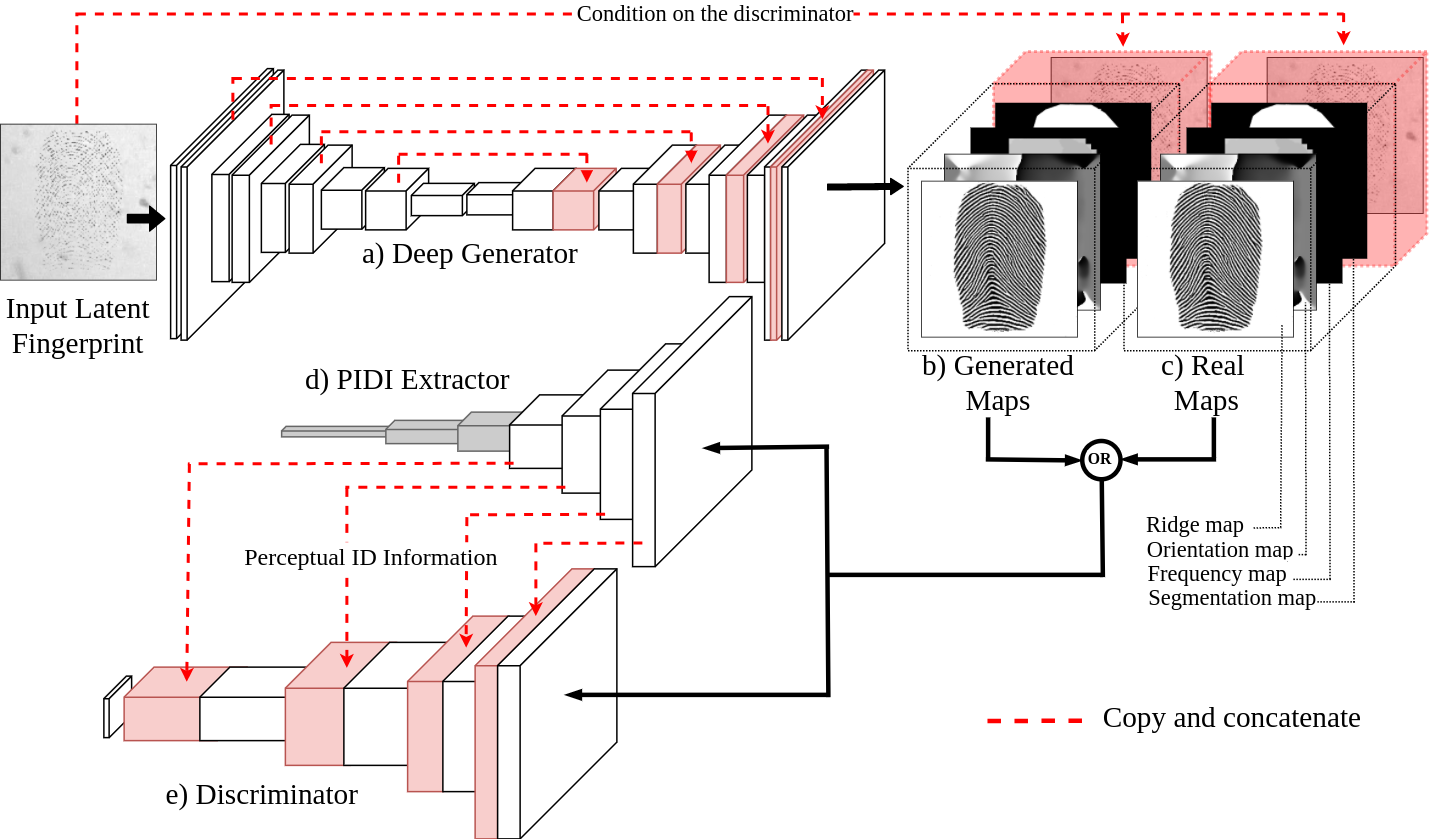}
\end{center}
   \caption{Complete diagram of the model. a) The deep generator takes the input latent fingerprints and generates a ridge, frequency, orientation and segmentation map simultaneously. b) Generated maps are concatenated with the input latent fingerprint to provide a condition for the discriminator. c) Real maps are extracted from the original clean fingerprints that were distorted to provide synthetic latent samples. During the training phase these maps are used to provide the supervision for the discriminator. d)One tower from a deep Siamese fingerprint verifier that was trained separately takes the generated or real maps and provides PIDI for the discriminator. e) The discriminator tries to distinguish between generated maps and the real maps using the combined ID and quality information from the generated maps.  }
\label{fig:diagram}
\end{figure*}

\subsection{cGAN for Latent Fingerprint Reconstruction}
The formulation and the network architecture of cGAN proposed by Isola et al. \cite{isola2016image} is a universal setup for image-to-image translation. However, processing a biometric image such as latent fingerprints is often more complicated than other image types due to the identification information that is embedded in the pattern of the image which should be preserved during the reconstruction process. For this purpose, we perform two modifications to the formulation and the network architecture to emphasize the ID of the input sample. 

First, to increase the quality of reconstruction, we force the network to generate four fingerprint maps $Y = G(x, \theta_g)$ for each latent fingerprint input as follows:
\begin{align}
\label{eq_generator}
Y = [y^R, y^F, y^O, y^S],
\end{align}
where, $Y$ is the new output of the generator, and consists of the ridge map ($y^R$), the orientation map ($y^O$), the frequency map ($y^F$) and the ridge segmentation map ($y^S$) of the input latent fingerprint. These four maps are concatenated depth-wise to form $Y$. Figure \ref{fig:diagram}(b) demonstrates the new output of the generator.    

Eq.~\ref{eq_l1} computes the direct reconstruction error. For the new outputs of the generator, the same formula is extended as:

\begin{align}
\label{eq_l1t}
l_{L1}(G) = \alpha_R ||{y^*}^R-y^R||_1 + \alpha_F ||{y^*}^F-y^F||_1 + \\ \alpha_O ||{y^*}^O-y^O||_1 + \alpha_S ||{y^*}^S-y^S||_1,
\end{align}
where, ${y^*}^R$, ${y^*}^F$, ${y^*}^O$ and ${y^*}^S$ are the ground truth values for the ridge, frequency, orientation and segmentation map respectively, and $\alpha_R$, $\alpha_F$, $\alpha_O$ and $\alpha_S$ are weights for scaling loss of reconstruction of each generated map. 
Since all maps except the ridge map have low-pass characteristic and may prevent the generator from predicting the ridge map with sufficient detail, we set $\alpha_R$ to 1.0 and all other coefficients ($\alpha_F$, $\alpha_O$ and $\alpha_S$) to 0.1. 

 Second, the generator should preserve the identification information embedded in minutiae and ridge patterns. To extract and preserve the identification information we developed a method that is inspired by perceptual loss \cite{dosovitskiy2016generating, gatys2016image, johnson2016perceptual, motiian2017few} and multi-level feature abstraction \cite{Soleymani2018multi, Soleymani2018generalized, kazemi2018facial, iranmanesh2018deep}. For this purpose, we separately trained a deep Siamese CNN as a fingerprint verifier. The trained model is used to extract the perceptual ID information (PIDI) of the generated maps. Extracted PIDI are output feature maps of the first four convolutional layers of the verifier module, and are concatenated to the corresponding layers of the discriminator to emphasize the ID information on the discriminator's decision. Figure \ref{fig:diagram}(d) shows how the output feature maps of the verifier contributes to the discriminator's decision making. Figure \ref{fig:discrim} shows the training criteria of the discriminator in more detail.      
 
\begin{table}[]
\centering
\caption{Architecture of the PIDI extractor. All layers except the last one comprise Convolution (C), Batch Normalization (B) \cite{ioffe2015batch} and ReLU (R). The output of the last layer is flatted after the convolution.  Convolution strides of all layers are two, and the spacial size of all kernels is $4\times4$. }
\label{tab:arch-verifier}
\begin{tabular}{|c|c|l|l|l|}
\hline
{L \#} & {Type} & \multicolumn{1}{c|}{K. Size} & \multicolumn{1}{c|}{Input Size} & \multicolumn{1}{c|}{Output Size} \\ \hline
1                              & C, B, R                        & $64$       & $256 {\times} 256{\times}4$     & $128{\times}128{\times}64$       \\ \hline
2                              & C, B, R                        & $128$      & $128{\times}128{\times}64$    & $64{\times}64{\times}128$        \\ \hline
3                              & C, B, R                        & $256$      & $64{\times}64{\times}128$       & $32{\times}32{\times}256$        \\ \hline
4                              & C, B, R                        & $512$      & $32{\times}32{\times}256$       & $16{\times}16{\times}512$        \\ \hline
5                              & C, B, R                        & $512$      & $16{\times}16{\times}512$       & $8{\times}8{\times}512$          \\ \hline
6                              & C, B, R                        & $512$       & $8 {\times} 8{\times}512$     & $4{\times}4{\times}512$       \\ \hline
7                              & C                        & $512$       & $4 {\times} 4{\times}512$     & $2048{\times}1$       \\ \hline
\end{tabular}
\end{table}


\subsection{Network Architecture}
The proposed model consists of three networks: a fingerprint PIDI extractor, a generator, and a discriminator. The fingerprint PIDI extractor is one tower from a deep Siamese \cite{chopra2005constrastiveloss} fingerprint verifier that is trained using a contrastive loss \cite{chopra2005constrastiveloss}. We fix all weights of the CNN tower, and use it to extract the PIDI of samples generated by the generator. Figure \ref{fig:diagram}(d) shows how the output of the first four layers of the PIDI extractor are fused into the discriminator to provide the identification information for the discriminator. Table \ref{tab:arch-verifier} details the architecture of the PIDI extractor.


The generator is a `U-net' \cite{ronneberger2015u, isola2016image} auto-encoder CNN. In the `U-net' architecture, some layers from the encoder are concatenated to layers of the decoder to keep the high frequency details of the input samples, and increase the quality of reconstruction in the decoder. Figure \ref{fig:diagram}(a) shows a diagram of the generator, and Table \ref{tab:arch-generator} details the structure of the generator. 

The discriminator is a deep CNN which maps the conditioned output of the generator with a size of $256\times256\times5$ to a discrimination matrix of size $16\times16\times1$. To force the discriminator to consider the ID information of the input samples, outputs of the first four layers of the PIDI extractor network are concatenated to the output of the corresponding layers of the discriminator. Table \ref{tab:arch-discrim} details the architecture of the discriminator network. The discriminator judges the output of the generator in a patch-wise manner. Each element of the discriminator's output map represents the discriminator's decision about a corresponding patch in the generator's output. 

The whole generative model was trained over 400 epochs, each epoch consisting of 7,812 iterations with a batch size = 64. Adam optimization method \cite{kingma2014} is used as the optimizer due to its fast convergence with beta = 0.5 and learning rate $= 10^{-4}$
.

\begin{table*}[]
\centering
\caption{Architecture of the discriminator. All layers except the last one comprise Convolution (C), Batch Normalization (B) \cite{ioffe2015batch} and ReLU (R). Non-linear function of the last layer is a Sigmoid (S). Convolution strides of the first four layers are two. The output of the first four layers of the verifier are concatenated to the corresponding four layers of the discriminator to enforce ID information to discriminator's decision making.}
\label{tab:arch-discrim}
\begin{tabular}{|c|c|c|l|l|l|l|l|}
\hline
Layer & Type & \multicolumn{1}{l|}{St.} & \multicolumn{1}{c|}{Kernel Size} & \multicolumn{1}{c|}{Input Size} & \multicolumn{1}{c|}{Output Size} & Concat. with & Final Output Size           \\ \hline
1                               & C, B, R                 & 2                        & $4\times4\times64$           & $256\times256\times5$       & $128\times128\times64$       & L 1          & $128\times128\times128$ \\ \hline
2                               & C, B, R                 & 2                        & $4\times4\times128$          & $128\times128\times128$         & $64\times64\times128$            & L 2          & $64\times64\times256$   \\ \hline
3                               & C, B, R                 & 2                        & $4\times4\times256$              & $64\times64 \times256$          & $32\times32\times256$            & L 3          & $32\times32\times512$   \\ \hline
4                               & C, B, R                 & 2                        & $4\times4\times512$              & $32\times32\times512$           & $16\times16\times512$            & L 4          & $16\times16\times1024$  \\ \hline
5                               & C, B, S                & 1                        & $4\times4\times1$                & $16\times16\times1024$          & $16\times16\times1$              & -            & $16\times16\times1$     \\ \hline
\end{tabular}
\end{table*}

\begin{table*}[]
\centering
\caption{Architecture of the generator. All layers except the last one comprise Convolution (C), Batch Normalization (B) \cite{ioffe2015batch} and ReLU (R). Non-linear function of the last layer is Sigmoid (S). Convolution strides of the first four layers are two, and the spacial size of all kernels is $4\times4$. }
\label{tab:arch-generator}
\begin{tabular}{|c|c|c|l|l|l|l|l|l|l|l|l|l|}
\hline
\#L & T                      & \multicolumn{1}{l|}{S} & \multicolumn{1}{c|}{\#K} & \multicolumn{1}{c|}{Input Size} & \multicolumn{1}{c|}{Output Size} & \#L & T & S & \#K & Input Size                  & Output Size                 & Cn \\ \hline
1   & C                      & 1                        & 64                       & $256{\times}256{\times}1$       & $256{\times}256{\times}64$       & 10  & C & 1   & 512 & $16{\times}16{\times}512$   & $16{\times}16{\times}512$   & -  \\ \hline
2   & C                      & 2                        & 128                      & $256{\times}256{\times}64$      & $128{\times}128{\times}128$      & 11  & D & 2   & 512 & $16{\times}16{\times}512$   & $32{\times}32{\times}512$   & L7 \\ \hline
3   & C                      & 1                        & 128                      & $128{\times}128{\times}128$     & $128{\times}128{\times}128$      & 12  & C & 1   & 512 & $32{\times}32{\times}1024$  & $32{\times}32{\times}512$   & -  \\ \hline
4   & C                      & 2                        & 256                      & $128{\times}128{\times}128$     & $64{\times}64{\times}256$        & 13  & D & 2   & 256 & $32{\times}32{\times}512$   & $64{\times}64{\times}256$   & L5 \\ \hline
5   & C                      & 1                        & 256                      & $64{\times}64{\times}256$       & $64{\times}64{\times}256$        & 14  & C & 1   & 256 & $64{\times}64{\times}512$   & $64{\times}64{\times}256$   & -  \\ \hline
6   & \multicolumn{1}{l|}{C} & 2                        & 512                      & $64{\times}64{\times}256$       & $32{\times}32{\times}512$        & 15  & D & 2   & 128 & $64{\times}64{\times}256$   & $128{\times}128{\times}128$ & L3 \\ \hline
7   & \multicolumn{1}{l|}{C} & 1                        & 512                      & $32{\times}32{\times}512$       & $32{\times}32{\times}512$        & 16  & C & 1   & 128 & $128{\times}128{\times}256$ & $128{\times}128{\times}128$ & -  \\ \hline
8   & \multicolumn{1}{l|}{C} & 2                        & 512                      & $32{\times}32{\times}512$       & $16{\times}16{\times}512$        & 17  & D & 2   & 64  & $128{\times}128{\times}128$ & $256{\times}256{\times}64$  & L1 \\ \hline
9   & \multicolumn{1}{l|}{C} & 1                        & 512                      & $16{\times}16{\times}512$       & $16{\times}16{\times}512$        & 18  & C & 1   & 4  & $256{\times}256{\times}128$ & $256{\times}256{\times}4$   & -  \\ \hline
\end{tabular}
\end{table*}

\begin{figure*}
\begin{center}
\includegraphics[scale=.4]{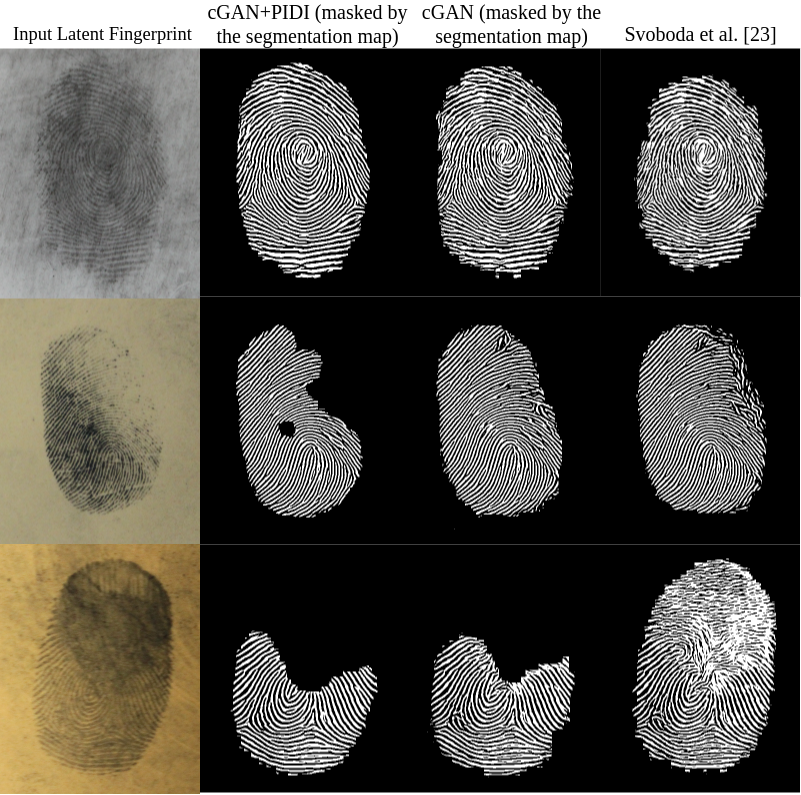}
\end{center}
   \caption{Examples of the fingerprint reconstructions on real latent fingerprints. For each fingerprint image, the corresponding latent sample and the reconstructed sample are demonstrated. Matching scores for the latent fingerprints and the reconstructed samples are calculated using VeriFinger. All samples are from the IIIT-Delhi latent fingerprint database \cite{sankaran2014latent}. 
}
\label{fig:latent_rec}
\end{figure*}

\subsection{Synthetic Dataset of Latent Fingerprints}
Training a generative model requires a large number of training samples. We synthetically generated a comprehensive database of latent fingerprints by distorting 5,000 clean fingerprints from the BioCOP 2013 \cite{biocop} database. For each clean fingerprint we generated 100 different latent samples. Our final database of latent fingerprints consists of 500,000 latent fingerprints and their corresponding clean samples. Examples of generated latent samples are shown in Figure \ref{fig:latent}.

\section{Experiments}
\label{experiment}
To evaluate the performance of the proposed latent fingerprint reconstruction technique, we conducted three different experiments on publicly available datasets of latent fingerprints. Unfortunately, NIST-SD27
\cite{nist27} is no longer available, so
the IIIT-Delhi Latent fingerprint \cite{sankaran2011matching} and IIIT-Delhi Multi Sensor Latent Fingerprint (MOLF) \cite{sankaran2015multisensor} databases were used to evaluate the proposed method.  
In all experiments, we used VeriFinger 7.0 SDK \cite{verifinger} and the NIST Biometrics Image Software (NBIS) \cite{watson2011} to match the reconstructed samples. In addition, to evaluate the role of the PIDI fusion which is developed to force the generator to preserve the ID information, we developed a second model which is exactly the same as our complete model but without the PIDI extractor module. Results for the complete model are named `cGAN+PIDI', and results for the second model are named `cGAN'. 

\subsection{Latent-to-sensor matching}
\begin{figure}
\begin{center}
\includegraphics[scale=0.58]{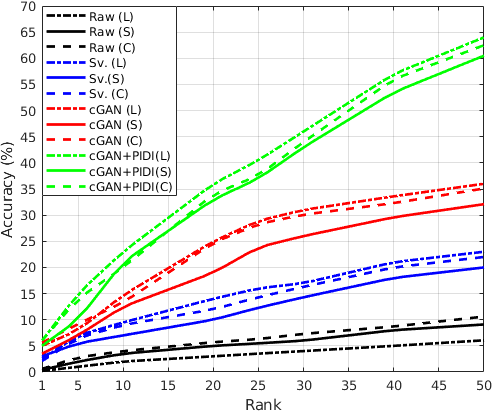}
\end{center}
\vspace{-10pt}
   \caption{CMC curves for the experiment of latent-to-sensor matching with NBIS. The reconstructed ridge maps were matched to fingerprints captured by the Lumidigm (L), Secugen (S) and Crossmatch (C) sensors.}
\label{fig:cmc-nbis}
\end{figure}

\begin{figure}
\begin{center}
\includegraphics[scale=0.58]{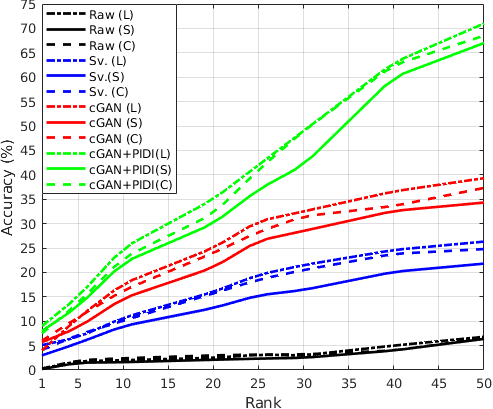}
\end{center}
\vspace{-10pt}
   \caption{CMC curves for the experiment of latent-to-sensor matching with VeriFinger. The reconstructed ridge maps were matched to fingerprints captured by the Lumidigm (L), Secugen (S) and Crossmatch (C) sensors.}
\label{fig:cmc-verif}
\end{figure}

For the first experiment, we used the IIIT-Delhi MOLF database. This database contains 19,200 fingerprint samples from 1000 classes (10 fingers of 100 individuals). For each ID, a set of latent fingerprint samples and the corresponding clean samples captured from three different commercial fingerprint scanners (Crossmatch, Secugen, Lumidigm) are available. As in the testing protocol established by Sankaran et al. \cite{sankaran2015multisensor}, the first and second fingerprint samples of each user captured by a sensor are selected as the gallery. The entire latent fingerprint database consisting of 4,400 samples used as the probe set. CMC curves for this experiment are shown in Figure \ref{fig:cmc-nbis} and Figure \ref{fig:cmc-verif}. Table \ref{nbis-table} and Table \ref{verif-table} show rank-25 and rank-50 accuracy for the latent-to-sensor matching experiment.

\subsection{Latent-to-latent matching}
\begin{figure}
\begin{center}
\includegraphics[scale=0.46]{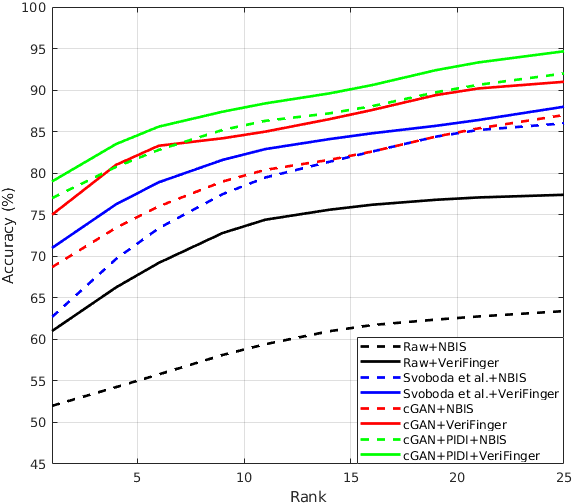}
\end{center}
\vspace{-10pt}
   \caption{CMC curves for the experiment of latent-to-latent matching using NBIS and VeriFinger.}
\label{fig:cmc-latent}
\end{figure}

For the second experiment, we evaluated the proposed method on the IIIT-Delhi latent fingerprint dataset which contains 1046 samples from all ten fingerprints recorded from 15 subjects.
The experimental setup is defined the same as \cite{sankaran2011matching} by randomly choosing 395 images as gallery and 520 samples as probes. 
The CMC curves for this experiment is shown in Figure \ref{fig:cmc-latent}. Rank-1, rank-10 and rank-25 results are also shown in Table \ref{verif-table1}. 

\subsection{Quality of the reconstructed fingerprints}
\begin{figure}
\begin{center}
\includegraphics[scale=0.65]{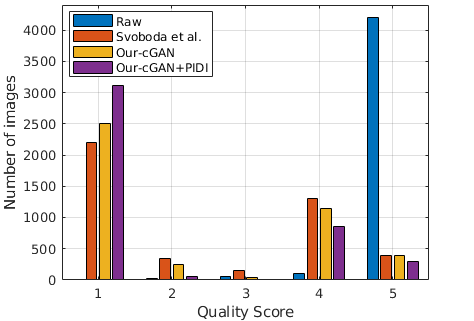}
\end{center}
\vspace{-10pt}
   \caption{Quality assessment of the reconstructed samples using NFIQ.}
\label{fig:score}
\end{figure}

Using the NFIQ utility from NBIS, the quality of reconstructed samples is measured to directly assess the performance of the reconstruction model. NFIQ assigns each fingerprint a numerical score from 1 (high quality) to 5 (low quality). Quality scores are computed for the reconstructed samples by our method and compared to score of both the raw latent fingerprints and those enhanced by the generative model developed by Svoboda et al. \cite{svoboda2017}. Figure \ref{fig:score} shows the quality scores of the reconstructed samples, and Fig. \ref{fig:latent_rec} shows three examples of the reconstructed fingerprints with different amount of distortion in the input sample.

\begin{table}[]
\centering
\caption{Latent-to-Sensor matching using NBIS on the MOLF database. }
\label{nbis-table}
\begin{tabular}{cccc}
\multicolumn{2}{c}{}                                                          & \multicolumn{2}{c}{Accuracy (\%)} \\ \cline{3-4} 
Sensor                    & Enhancement                                       & Rank-25         & Rank-50         \\ \hline
\multirow{4}{*}{Lumidigm} & Raw                                               & 3.52            & 6.06            \\ \cline{2-4} 
                          & Svoboda et al.\cite{svoboda2017} & 16.71           & 23.03           \\ \cline{2-4} 
                          & cGAN                                              & 28.82           & 36.07           \\ \cline{2-4} 
                          & cGAN+PIDI                                         & \textbf{40.52}           & \textbf{64.80}           \\ \hline
Secugen                   & Raw                                               & 5.48            & 9.19            \\ \cline{2-4} 
                          & Svoboda et al.\cite{svoboda2017} & 12.33           & 20.47           \\ \cline{2-4} 
                          & cGAN                                              & 23.68           & 32.16           \\ \cline{2-4} 
\multicolumn{1}{l}{}      & cGAN+PIDI                                         & \textbf{37.67}           & \textbf{60.58}           \\ \hline
CrossMatch                & Raw                                               & 6.01            & 10.64           \\ \cline{2-4} 
                          & Svoboda et al.\cite{svoboda2017} & 14.39           & 22.73           \\ \cline{2-4} 
                          & cGAN                                              & 28.37           & 35.23           \\ \cline{2-4} 
\multicolumn{1}{l}{}      & cGAN+PIDI                                         & \textbf{37.61}           & \textbf{62.55}           \\ \hline
\end{tabular}

\end{table}


\begin{table}[]
\centering
\vspace{2mm}
\caption{Latent-to-Sensor matching using VeriFinger on the MOLF database.}
\label{verif-table}
\begin{tabular}{cccc}
\multicolumn{2}{c}{}                                                            & \multicolumn{2}{c}{Accuracy (\%)} \\ \cline{3-4} 
Sensor                      & Enhancement                                       & Rank-25         & Rank-50         \\ \hline
\multirow{4}{*}{Lumidigm}   & Raw                                               & 3.13            & 6.80            \\ \cline{2-4} 
                            & Svoboda et al.\cite{svoboda2017} & 19.51           & 26.24           \\ \cline{2-4} 
                            & cGAN                                              & 30.47           & 39.38           \\ \cline{2-4} 
                            & cGAN+PIDI                                         & \textbf{42.04}           & \textbf{70.89}           \\ \hline
\multirow{4}{*}{Secugen}    & Raw                                               & 2.33            & 6.37            \\ \cline{2-4} 
                            & Svoboda et al.\cite{svoboda2017} & 15.23           & 21.81           \\ \cline{2-4} 
                            & cGAN                                              & 26.44           & 34.32           \\ \cline{2-4} 
                            & cGAN+PIDI                                         & \textbf{37.14}           & \textbf{66.11}           \\ \hline
\multirow{4}{*}{CrossMatch} & Raw                                               & 3.17            & 6.51            \\ \cline{2-4} 
                            & Svoboda et al.\cite{svoboda2017} & 18.34           & 24.78           \\ \cline{2-4} 
                            & cGAN                                              & 28.30           & 37.31           \\ \cline{2-4} 
                            & cGAN+PIDI                                         & \textbf{41.27}           & \textbf{68.61}           \\ \cline{2-4} 
\end{tabular}

\end{table}

\begin{table}[]
\centering
\caption{Rank-1, rank-10 and rank-25 results for the experiment of latent-to-latent matching on the IIIT-Delhi latent database.}
\label{verif-table1}
\begin{tabular}{cccc}
                                                & \multicolumn{3}{c}{Accuracy (\%)} \\ \cline{2-4} 
Enhancement                                     & Rank-1    & Rank-10   & Rank-25   \\ \hline
Raw  +NBIS                                      & 52.31     & 58.90     & 63.42     \\ \hline
\cite{svoboda2017} +NBIS       & 62.69     & 78.85     & 86.12     \\ \hline
cGAN +NBIS                                      & 68.69     & 79.85     & 87.23     \\ \hline
cGAN+PIDI+NBIS                                  & 77.16     & 86.04     & 92.10     \\ \hline
Raw +VeriFinger                                 & 61.02     & 74.00     & 77.44     \\ \hline
\cite{svoboda2017} +VeriFinger & 71.04     & 82.56     & 88.28     \\ \hline
cGAN +VeriFinger                                & 74.92     & 86.51     & 91.11     \\ \hline
cGAN+PIDI+VeriFinger                            & \textbf{79.23}     & \textbf{88.02}     & \textbf{94.67}     \\ \hline
\end{tabular}
\end{table}

\section{Conclusion}
\label{conclusion}

Recognizing latent fingerprint samples is a challenging problem for identification systems since a latent fingerprint image can be `noisy' with a large portions of the fingerprint missing, leading to a lower amount of ridge information compared to normal fingerprints. Following the successful outcomes of exploiting deep generative models for the traditional image processing problems, such as denoising, inpainting, and image to image translations, we propose a deep latent fingerprint reconstruction model based on conditional generative adversarial networks. We applied two modifications to the cGAN formulation and network architecture to adapt it for the task of latent fingerprint reconstruction. Generated ridge maps using GAN models often contain random ridge patterns for severely distorted areas of the input fingerprint. Main generator of our model is forced to generate three extra fingerprint maps. One of these maps is the ridge segmentation map which shows the reliability of the corresponding ridge map. 

Opposed to the previous works in the literature, the proposed network directly translates the input latent fingerprints to the clean binary ridge maps by predicting the missing ridge information. Incorporating a discriminator network which measures both quality and PIDI of the reconstructed samples simultaneously, along with the generation process in the training phase, increases the quality of the generated samples directly without a need to define multiple complex loss functions for minimizing the similarity between the generated ridge patterns and the ground truth maps.

The proposed method successfully reconstructed latent fingerprints from the IIIT-Delhi latent and IIIT-Delhi MOLF databases in different experimental setups of latent-to-sensor and latent-to-latent matching. We achieved rank-50 accuracy of 70.89\% for the latent-to-sensor matching on the IIIT-Delhi MOLF database. For the latent-to-latent matching we achieved rank-10 accuracy of 88.02\%. Although the best results in both experiments were obtained when VeriFinger was used as the matcher, NBIS matching algorithm also resulted in high matching accuracy.   

In addition, measuring the quality of reconstructed fingerprints using NFIQ shows that the generated fingerprints are significantly enhanced compared to the raw latent samples.

For future work it is desired to directly extract minutiae from the latent input fingerprints. On the other hand, increasing the size of the synthetic latent database by introducing more complex distortions is another future direction for this work.

\begin{center}
ACKNOWLEDGEMENT
\end{center}

This work is based upon a work supported by the Center
for Identification Technology Research and the National
Science Foundation under Grant \#1650474.

{\small
\bibliographystyle{ieee}
\bibliography{egbib}
}

\end{document}